\newif\ifcomments  
\commentstrue

\documentclass[table]{article}
\usepackage{spconf,amsmath,graphicx}
\usepackage[dvipsnames]{xcolor}
\usepackage{highlight2}
\usepackage{comment}
\usepackage{hyperref}
\usepackage{url}
\usepackage{enumitem}
\usepackage{booktabs}
\usepackage{mathtools}
\usepackage{dirtytalk}
\usepackage{multirow}

\title{Multilingual and Fully Non-Autoregressive ASR with Large Language Model Fusion: A Comprehensive Study}
\name{\begin{tabular}{c}W. Ronny Huang, Cyril Allauzen, Tongzhou Chen, Kilol Gupta, Ke Hu, James Qin, Yu Zhang, \\ Yongqiang Wang, Shuo-Yiin Chang, Tara N. Sainath\end{tabular}}
\address{Google Research}

\begin{document}
\maketitle
\begin{abstract}

In the era of large models, the autoregressive nature of decoding often results in latency serving as a significant bottleneck. We propose a non-autoregressive LM-fused ASR system that effectively leverages the parallelization capabilities of accelerator hardware. Our approach combines the Universal Speech Model (USM) and the PaLM 2 language model in per-segment scoring mode, achieving an average relative WER improvement across all languages of 10.8\% on FLEURS and 3.6\% on YouTube captioning. Furthermore, our comprehensive ablation study analyzes key parameters such as LLM size, context length, vocabulary size, fusion methodology. For instance, we explore the impact of LLM size ranging from 128M to 340B parameters on ASR performance. This study provides valuable insights into the factors influencing the effectiveness of practical large-scale LM-fused speech recognition systems.

\end{abstract}
\begin{keywords}
large language model, multilingual speech recognition
\end{keywords}

\section{Introduction}

Large-scale models have showcased impressive results across applications, especially in multilingual speech models \cite{pratap2020massively,li2021scaling,zhang2022bigssl,radford2023robust,chen2023improving} and multilingual large language models (LLMs) \cite{shoeybi2019megatron,brown2020language,touvron2023llama}. The next evolution is the fusion of these large multilingual models to enhance accuracy. However, their size brings latency issues, notably in autoregressive decoding, as seen with shallow fusion \cite{hu2023massively}.

Applications like voice assistants and live captioning face challenges from these models' latency. Recognizing this concern, our study presents a non-autoregressive LM-fused ASR system. This approach streams utterances at 8-second intervals using the Universal Speech Model (USM) \cite{zhang2023google} and the PaLM 2 language model \cite{anil2023palm}. 
Both the USM and PaLM 2 process each 8-second chunk with full audio context, leading to enhanced accuracy,
and they are also able process it quickly to by parallelizing across the sequence for smooth user experience.
In particular, hypotheses are generated non-autoregressively by attaching a CTC decoder to the USM, and the hypotheses are scored non-autoregressively by the LM by passing in the entire hypothesis sequence in teacher forcing mode \cite{salazar2019masked}.

Our methodology yields an average double-digit gain of 10.8\% on relative word error rate (WER) on the public multilingual FLEURS testset. On a challenging internal YouTube captioning testset, we observe an average gain of 3.6\%, across all available languages. Our study further delves into factors like LM size (up to 340B), vocabulary, context length, segmentation, n-best list size, and scoring method. For instance, we find that though LLMs display emergent behaviors with increasing size \cite{wei2022emergent}, their impact on WER is more muted---but larger models can reduce the sensitivity to fusion weight.

\section{Related work}
Leveraging large language models to enhance ASR systems has emerged as a natural and promising direction, leading to significant advancements in recent research. Several studies have focused on integrating LLMs with ASR models, exploiting their vast linguistic knowledge and contextual understanding.
\cite{chen2023large} combined T5 \cite{raffel2020} and PaLM 1 \cite{chowdhery2022palm} with a Conformer RNN-T model, enhancing ASR performance for English and code-switched data. Our work builds on this by using LLMs for large-scale ASR models with a non-autoregressive CTC head.

\cite{hu2023massively} improved recognition of rare words in short voice queries using shallow fusion. In contrast, our focus lies in long-form tasks like YouTube captioning, where LLMs' contextual understanding is crucial. Given the limitations of shallow fusion for long tasks and mismatched vocabularies, we explore scoring as a more efficient fusion technique.

\cite{yu2022non} integrated a Listen Attend Spell Once model \cite{bai2021fast} with BERT \cite{Devlin2018} using n-best scoring. We expand upon this idea by deploying larger-scale, multilingual models.
Other research, such as \cite{Anjuli18}, has explored various fusion methods between LLMs and ASR systems. Their surprising finding was the comparable performance of simple shallow fusion to deeper techniques. Similarly, \cite{li2023prompting} used two fusion methods with LLaMa LLM and found minimal difference in WER outcomes, supporting the conclusions of \cite{Anjuli18}.

Building on these findings, our study emphasizes scoring as the fusion method, aiming for a practical and scalable ASR+LLM solution suitable for real-world applications.

\section{Method}
\label{sec:method}

\subsection{Speech Model}
We employ the Universal Speech Model (USM) \cite{zhang2023google}, a 2 billion parameter Conformer \cite{gulati2020conformer}, with 32 layers and a model dimension of 1536, for ASR hypotheses. The vocabulary comprises 16384 wordpieces, and a CTC decoder ensures non-autoregressive, parallel inference.
For training the USM, a multitask approach is used. It's trained on over 12 million hours of unlabeled audio and 28 billion sentences of text data, along with 110 thousand hours of supervised and 100 thousand hours of semi-supervised audio. All datasets are multilingual.
The USM features chunk-wise bi-directional attention, enabling accurate long-form audio modeling with 30 second segment during training. Unlike traditional audio-based chunking, this approach maintains continuous state throughout, allowing for streaming results every 8 seconds, enhancing user experience.

\subsection{Language Model}
We utilize the PaLM 2 language model \cite{anil2023palm} to score the ASR hypotheses. Trained on varied data sources like web documents and books, it uses a 256k wordpiece vocabulary. PaLM 2 surpasses its predecessor, PaLM 1 \cite{chowdhery2022palm}, via enhanced training, architecture improvements, and extended context length, showcasing superior performance in natural language tasks. We assess its capability in ASR scoring using the pre-trained variant and apply prefix LM scoring mode \cite{wang2022language}, prompting the model with a fixed prefix (top hypotheses from previous segments) and scoring several suffix hypotheses (different hypotheses for current segment).

\subsection{Long-form Inference}
To process long-form audio without memory constraints, we employ a streaming framework, processing the audio frame by frame. Using the USM's chunk-wise attention, we encode 8-second chunks as soon as the audio is available and relay them to the CTC decoder. These CTC probabilities form a confusion network lattice encoding possible wordpieces. Given the independence of each encoded frame, the wordpiece distributions are also independent of one another. Consequently, the lattice holds hypotheses that exponentially grow with length, making it challenging to score all of them with an LLM.

\begin{figure*}[]
  \centering
  \vspace{-14pt}
  \includegraphics[width=0.70\textwidth]{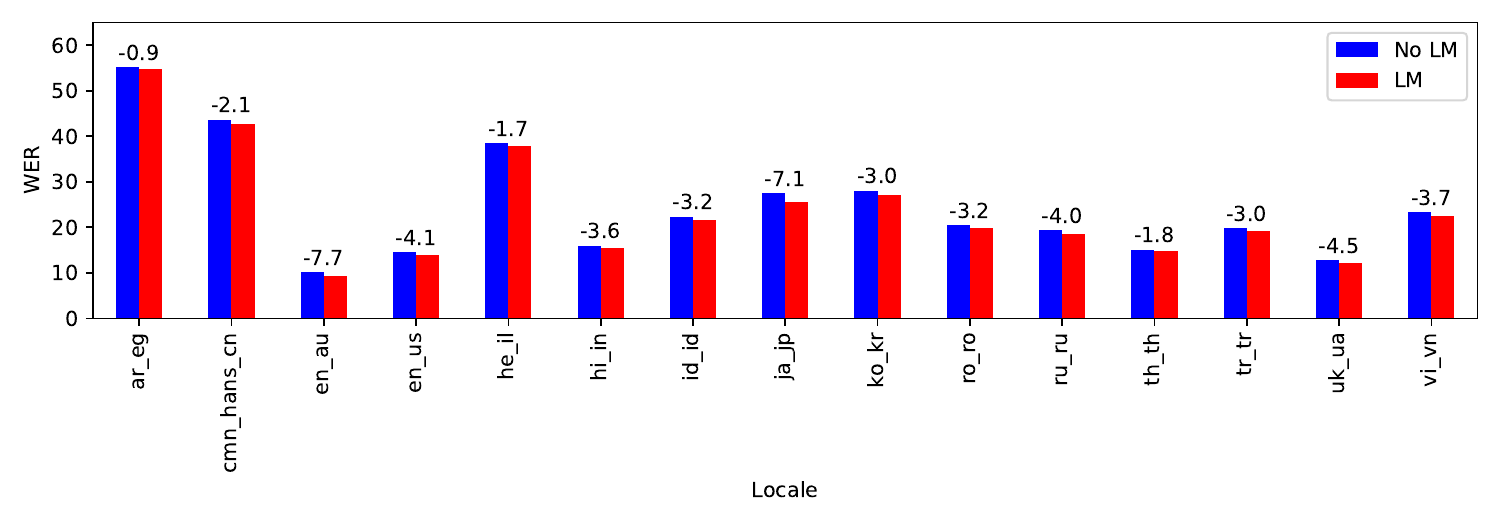} \\
  \includegraphics[width=0.75\textwidth]{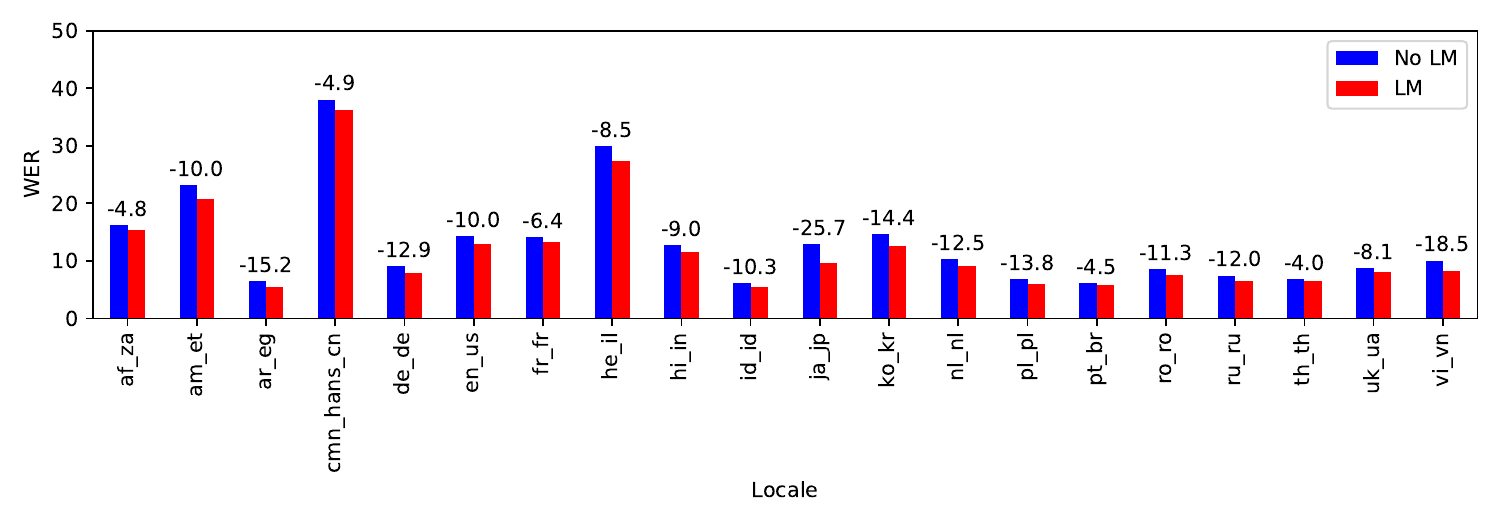} \\
  \vspace{-10pt}
  \caption{Results on all languages. Text indicates relative WER change. Top: Youtube Captions testset. Bottom: FLEURS testset.}
  \vspace{-10pt}
  \label{fig:multilang}
\end{figure*}

\subsection{LM Scoring}
To handle the exponential growth, we break the lattice into 8-second segments, scoring only current segment hypotheses. To maintain context, we use the concatenated top hypothesis from the previous two segments as the LM prefix. This iterative process updates every 8 seconds, ensuring the prefix is always the top hypotheses from the last two segments. To derive suffixes for the LM, we pick the best N hypotheses from the lattice and batch them as suffix sequences for scoring. The combined LM and ASR scores given the audio $\textbf{x}$ and the hypothesis $\textbf{y}$ are then represented by
\vspace{-5px}
\begin{equation}
    \log p_{final}(\textbf{y}|\textbf{x}) = \log p_{asr}(\textbf{y}|\textbf{x}) + \lambda\cdot \log p_{lm}(\textbf{y}),
\end{equation}
\noindent with the leading hypothesis chosen as the final transcript.
Per-segment scoring is parallelizable, non-autoregressive, and updates in streaming mode every 8 seconds as one speaks.

\section{Evaluation}
\label{sec:evaluation}

Throughout our evaluations, we will use the following setup unless otherwise mentioned.
\vspace{-5pt}
\begin{itemize}[leftmargin=*, itemsep=-0.4em]
    \item US English locale
    \item 1 billion parameter variant of PaLM 2
    \item LM scoring weight of 0.3 (optimized according to the settings list here)
    \item Context length of 2 prior segments
    \item N-best list size of 16
    \item Youtube Captions testset, described in next section.
\end{itemize}

\subsection{Testsets}
YouTube videos span diverse categories, making them suitable for our LM scoring study. We evaluate on the commonly used YouTube captioning testset, YT\_LONG, encompassing video-on-demand content in multiple languages \cite{Soltau2017,chiu2019comparison,chiu2021rnn,huang2022e2e,chen2023large}. For US English, it includes 77 videos totaling 22.2 hours, with a median utterance length of 14.8 minutes. Other languages average 61 videos and 17 hours.
We also analyze results on the FLEURS testset \cite{conneau2023fleurs}, containing 600-900 utterances per language, with US English having 647. The total duration across languages is 283 hours.

\subsection{Results on All Languages}
Results for all languages are presented in Figure \ref{fig:multilang}. We tested LM scoring weight $\lambda$ at four values: \{0.15, 0.30, 0.45, 0.60\}.
On Youtube (Figure \ref{fig:multilang} (top)), PaLM 2 integration reduces WER by 4.1\% for en\_us, averaging 3.6\% across 15 languages.
The FLEURS testset (Figure \ref{fig:multilang} (bottom)) shows a more pronounced improvement: 9.7\% for en\_us and 10.8\% on average. No language showed regression.

\subsection{Dependence on LM Size}
\label{sec:lmsize}

\begin{figure}[]
  \centering
  \vspace{-6pt}
  \includegraphics[width=0.8\linewidth]{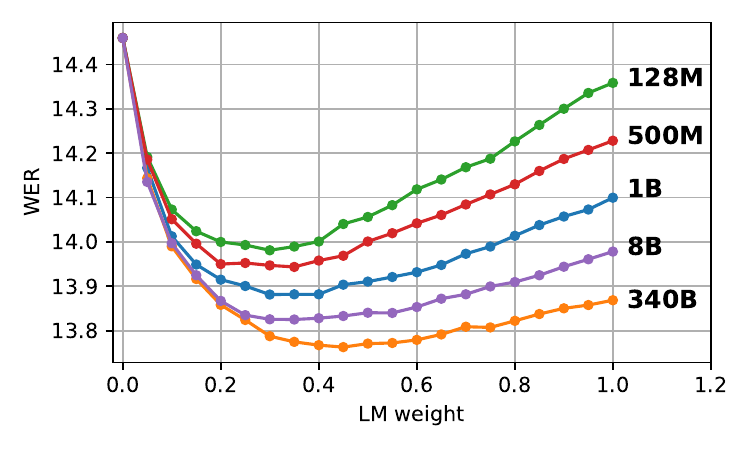}
  \vspace{-14pt}
  \caption{Dependence of various PaLM 2 models on the LM scoring weight.}
  \label{fig:lmsize}
\end{figure}

Large language models demonstrate emergent abilities with increased size, data, and compute \cite{wei2022emergent,bubeck2023sparks}. This study examines ASR hypothesis scoring using different PaLM 2 scale variants. Results in Table \ref{tab:lmsize} indicate that while WER improves with larger models (E2-E6), the gains might not offset the growing inference costs. Additionally, optimal LM scoring weight increases with model size, shifting from 0.25 for a 128M LM to 0.45 for a 340B LM (Figure \ref{fig:lmsize}). Larger models show decreased WER sensitivity to LM weight changes. This suggests that smaller models require cautious weighting, while larger models, with their improved accuracy, can afford more weight without risking incorrect hypothesis selection.

\begin{table}[]
\centering
\caption{Dependence on PaLM 2 model size.}
\label{tab:lmsize}
\vspace{5pt}
\begin{tabular}{l|ll}
\toprule
LM size & WER & relative   \\
\midrule
B1: No LM & 14.48 & -  \\
E2: 128M &  13.98 & \textcolor{red}{-3.4\%}    \\
E3: 500M &  13.94 & \textcolor{red}{-3.7\%}    \\
E4: 1B &    13.88 & \textcolor{red}{-4.1\%}    \\
E5: 8B &    13.83 & \textcolor{red}{-4.5\%}    \\
E6: 340B &  13.76 & \textcolor{red}{-5.0\%}    \\
\bottomrule
\end{tabular}
\end{table}
\vspace{-5pt}

\subsection{Dependence on Context Length}
\begin{figure}[]
  \centering
  \includegraphics[width=0.8\linewidth]{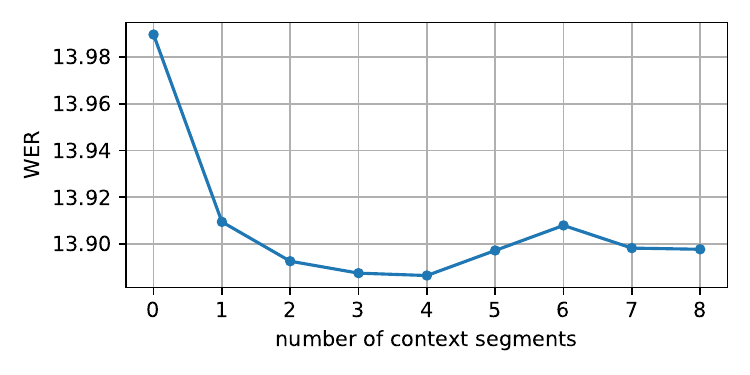}
  \vspace{-14pt}
  \caption{Dependence on number of context segments.}
  \label{fig:ctxlen}
\end{figure}

By adjusting the number of segments from previous history to utilize as a prompt, we can coarsely control the LM's context length. Figure \ref{fig:ctxlen} indicates that concatenating 4 context segments, or 32 seconds of decoded text, is optimal. Including more than 4 segments slightly reduces performance, possibly due to the LM scoring weight being optimized at 2 segments. The results suggest that using about 32 seconds or approximately 50 words of context improves ASR. However, adding more context after this offers limited benefit, differing from many NLP tasks where longer contexts are essential.

\subsection{Dependence on Vocabulary Size}

PaLM 2 has a vocabulary of 256k tokens, optimized for natural language generation. A benefit of per-segment scoring is that it allows handling of mismatched vocabularies between ASR and LLM models through re-tokenization. In the 1-billion-parameter PaLM 2, embedding and softmax layers take up about one third of its parameters. We tested a reduced 32k token vocabulary for PaLM 2 to decrease computational cost. Results in Table \ref{tab:vocab} show minimal performance degradation with the smaller vocabulary. Thus, using a smaller vocabulary can save computation while retaining strong performance.

\begin{table}[]
\centering
\caption{Dependence on LM vocabulary size.}
\label{tab:vocab}
\vspace{5pt}
\begin{tabular}{l|l}
\toprule
LM vocabulary size & WER   \\
\midrule
B2: 256k      & 13.88 \\
E7: 32k           & 13.89 \\
\bottomrule
\end{tabular}
\end{table}

\subsection{Dependence on Segmentation}

Segmentation determines the frequency of lattice scoring with the LLM, influencing user experience and transcription quality \cite{huang2022e2e,huang2022e2esegmentation,huang2023semantic}. We evaluated fixed segmentation lengths and a voice activity detector (VAD) segmenter \cite{zazo2016feature}. While VAD avoids cutting words, it yields inconsistent segment lengths potentially affecting user experience when used in a per-segment streaming scenario. The median length of VAD segments is around 5 seconds.

\begin{figure}[]
  \centering
  \includegraphics[width=0.8\linewidth]{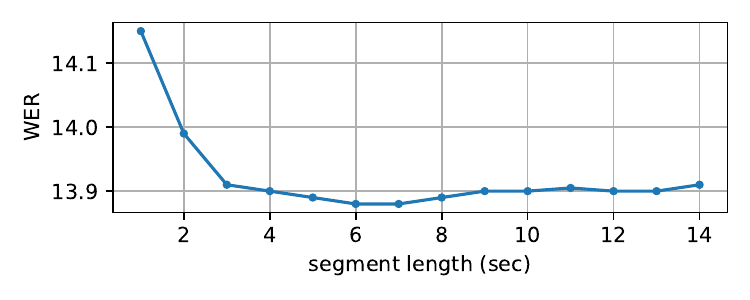}
  \vspace{-14pt}
  \caption{Dependence on the segment length.}
  \label{fig:seglen}
\end{figure}

Results in \ref{tab:seglen} show fixed-length (B3) surpasses VAD (E8), opposing findings in \cite{huang2022e2esegmentation}. This is due to differences in the model; \cite{huang2022e2esegmentation} uses RNN-T which discards most hypotheses upon segmentation, while our CTC model doesn't retain decoder states, making it more robust to premature segmentation.
\ref{fig:seglen} shows WER stability beyond 3 seconds. This is evidence that, in contrast to RNN-T, CTC remains unaffected by word truncation, thanks to its non-dependent confusion network structure.

\begin{table}[]
\centering
\vspace{-10pt}
\caption{Dependence on segmenter.}
\label{tab:seglen}
\vspace{5pt}
\begin{tabular}{l|l}
\toprule
Segmentation & WER   \\
\midrule
B3: Fixed 8 sec  & 13.88 \\
E8: VAD          & 13.98 \\
\bottomrule
\end{tabular}
\end{table}

\subsection{Dependence on Number of Hypotheses}
\begin{figure}[]
  \centering
  \vspace{-6pt}
  \includegraphics[width=0.83\linewidth]{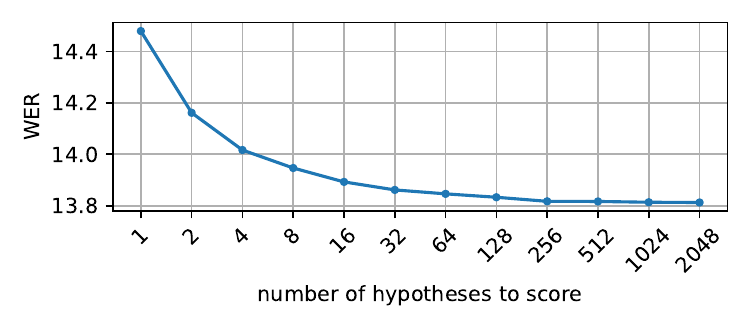}
  \vspace{-14pt}
  \caption{Dependence on n-best list size.}
  \label{fig:nbest}
\end{figure}

The number of paths in a lattice increases with segment length, but computational bounds limit the number of hypotheses scored. Figure \ref{fig:nbest} presents a study on the n-best list size, which denotes the scored hypotheses per segment. Performance improves as the list expands, plateauing at about 1024. This growth suggests the lattice's high density, allowing the LLM to continue improving the transcription quality up to computational constraints.

\subsection{Comparison to Shallow Fusion}

Our exploration has mainly considered per-segment LM scoring. An alternative is per-frame scoring or shallow fusion, which is computationally heavier due to its autoregressive decoding and frequent LM invocation.

In per-frame scoring, the LM acts on each frame, usually pre-pruning, amplifying computational load. When applied post-pruning, per-frame scoring requires forward propagations of $N_{frames}\times N_{hyps}$, while per-segment scoring demands $N_{tokens}\times N_{hyps}$ propagations. This makes per-frame about $N_{frames}/N_{tokens}$ times costlier. On our YouTube testset, this ratio is 4.
We apply a blank pruning strategy where frames with a blank probability above 0.9 are skipped. This largely diminishes the factor-of-4 cost difference.

Performance comparisons in Table \ref{tab:shallowfusion} show per-frame scoring (E9) at 13.70 (-5.4\% relative to no LM) outperforms per-segment scoring (B4) at 13.88 (-4.1\% relative). Shallow fusion shines in non-latency-critical scenarios with matched vocabularies. For per-frame scoring, we retrained the ASR model with PaLM 2's vocabulary.

\begin{table}[]
\centering
\caption{Comparison between per-segment to per-frame scoring (i.e. shallow fusion).}
\label{tab:shallowfusion}
\vspace{5pt}
\begin{tabular}{l|l}
\toprule
Vocabulary size & WER   \\
\midrule
B4: Per-segment scoring (scoring)      & 13.88 \\
E9: Per-frame scoring (shallow fusion) & 13.70 \\
\bottomrule
\end{tabular}
\end{table}
\section{Conclusion}
We developed a deployable ASR system using large-scale multilingual models, emphasizing practicality. By adopting a non-autoregressive system with CTC and per-segment LM scoring, we enhanced performance across languages for YouTube captions and FLEURS. Our study also provided insights into system parameters' effects on ASR efficacy.

\newpage
\small

\bibliographystyle{IEEEtrans}
\small
\bibliography{refs}

\end{document}